\documentclass[conference]{IEEEtran}
\IEEEoverridecommandlockouts
\usepackage{cite}
\usepackage{amsmath,amssymb,amsfonts}
\usepackage{algorithmic}
\usepackage{graphicx}
\usepackage{textcomp}
\usepackage{xcolor}
\usepackage{placeins} 

\def\BibTeX{{\rm B\kern-.05em{\sc i\kern-.025em b}\kern-.08em
    T\kern-.1667em\lower.7ex\hbox{E}\kern-.125emX}}

\begin{document}
\newpage
\thispagestyle{empty} 
\onecolumn 
\begin{center}
    \vspace*{5cm} 
    \textbf{© 2024 IEEE. Personal use of this material is permitted. Permission from IEEE must be obtained for all other uses,\\
    in any current or future media, including reprinting/republishing this material for advertising or promotional purposes, creating new collective works, for resale or redistribution to servers or lists, or reuse of any copyrighted component of this work in other works.}\\[2cm] 
    \textbf{Submitted and Presented at the IEEE International Conference on Innovative Engineering Sciences and Technological Research (ICIESTR-2024)}
\end{center}
\vspace*{\fill} 
\twocolumn 

\newpage
\title{Enhancing Object Detection with Hybrid dataset in Manufacturing Environments: Comparing Federated Learning to Conventional Techniques\\
\thanks{This work was funded by the Carl Zeiss Stiftung, Germany under the Sustainable Embedded AI project (P2021-02-009). \\
\\979-8-3503-4863-7/  24/ \$31.00~\copyright2024 IEEE}
}
\author{\IEEEauthorblockN{Vinit Hegiste}
\IEEEauthorblockA{\textit{Chair of Machine Tools and Control Systems} \\
\textit{RPTU Kaierslautern-Landau}\\
Kaiserslautern, Germany \\
vinit.hegiste@rptu.de}
\and
\IEEEauthorblockN{Snehal Walunj}
\IEEEauthorblockA{\textit{Innovative Factory Systems (IFS)} \\
\textit{German Research Center for Artificial Intelligence (DFKI)}\\
Kaiserslautern, Germany \\
snehal.walunj@dfki.de}
\and
\IEEEauthorblockN{Jibinraj Antony}
\IEEEauthorblockA{\textit{Innovative Factory Systems (IFS)} \\
\textit{German Research Center for Artificial Intelligence (DFKI)}\\
Kaiserslautern, Germany \\
jibinraj.antony@dfki.de}
\and
\IEEEauthorblockN{Tatjana Legler}
\IEEEauthorblockA{\textit{Chair of Machine Tools and Control Systems} \\
\textit{RPTU Kaierslautern-Landau}\\
Kaiserslautern, Germany \\
tatjana.legler@rptu.de}
\and
\IEEEauthorblockN{Martin Ruskowski}
\IEEEauthorblockA{\centerline{\textit{Innovative Factory Systems (IFS)}} \\
\textit{German Research Center for Artificial Intelligence (DFKI)}\\
Kaiserslautern, Germany \\
martin.ruskowski@dfki.de}
}
\maketitle

\begin{abstract}
Federated Learning (FL) has garnered significant attention in manufacturing for its robust model development and privacy-preserving capabilities. This paper contributes to research focused on the robustness of FL models in object detection, hereby presenting a comparative study with conventional techniques using a hybrid dataset for small object detection. Our findings demonstrate the superior performance of FL over centralized training models and different deep learning techniques when tested on test data recorded in a different environment with a variety of object viewpoints, lighting conditions, cluttered backgrounds, etc. These results highlight the potential of FL in achieving robust global models that perform efficiently even in unseen environments. The study provides valuable insights for deploying resilient object detection models in manufacturing environments.
\end{abstract}


\section{Introduction}
Federated learning (FL) has gained significance in the manufacturing domain owing to its ability to produce robust models while preserving data privacy, which is critical for companies to avoid data leaks. Recent research efforts, such as \cite{DasaradharamiReddy.2023}, \cite{Lyu.342020}, \cite{Xu.7152022}, \cite{Arazzi.3623114}, have focused on the privacy-preserving aspects, aiming to train models that match the performance of those trained using centralized datasets.
FL not only contributes to data-privacy protection but also yields robust models by addressing bias and handling repetitive data samples, as highlighted in \cite{Hegiste.92320229242022}, \cite{Hegiste.}, and \cite{Tayebi.2023}. Additionally, federated learning techniques demonstrate the ability to generate robust models that outperform centrally trained models during testing in unseen environments, as discussed in \cite{Hegiste.b}. Despite the evident advantages, there is limited coverage of research papers or scientific work focusing on this robust aspect of federated learning models.
Therefore, this paper aims to compare federated learning for object detection with conventional object detection techniques and other deep learning methods using a hybrid dataset specifically for small object detection.
In the realm of computer vision, particularly object detection, small objects are defined as those whose size is relatively minuscule in comparison to the complete image frame, typically less than 32*32 pixels \cite{antony2024enhancing}. Detection of small objects poses challenges due to limited features, lack of spatial information, lower context in terms of background, noise clutter, occlusion, and intra-class variability \cite{antony2024enhancing}.

The motivation extends beyond detecting small objects to include a comprehensive comparison of the efficiency of federated learning models. The hybrid dataset strategy involves using a synthetic dataset as one client and a real dataset as another client. The hybrid dataset is also subjected to training using the conventional YOLOv5 algorithm, incorporating techniques such as transfer learning and fine-tune learning with the synthetic dataset, followed by training with the real dataset. This comprehensive training approach enables further comparison with the federated learning model.
Another algorithm explored in this study is the YOLO ensemble technique, where models trained with synthetic and real datasets are ensembled to produce the final output.

\FloatBarrier
\section{Literature Review}

The task of detecting small objects within the field of computer vision poses a significant research challenge, due to the variability in the effectiveness of detection methods depending on specific applications, the nature of the data used, and the available computational resources. Object detection is crucial for a wide range of applications, such as autonomous driving technologies, industrial automation systems \cite{precup2023recognising}, extended reality experiences, surveillance systems \cite{yundong2020multi}, and analysis of aerial imagery. 
There has been a notable surge in the development and utilization of synthetic datasets within computer vision research \cite{borkman2021unity}, \cite{Acar2021-feddyn}, \cite{exploitingMultimodalSynthetic}, driven by the cost-effectiveness of synthetic data generation methods compared to traditional data collection techniques. These synthetic datasets are increasingly being adopted across various sectors, although some applications still rely on real datasets to minimize the domain gap between training and real-world data.

Small objects, defined as those with dimensions smaller than 32*32 pixels, present unique challenges for object detection models due to their limited feature sets, lower resolution, and consequent reduction in contextual information available for learning \cite{liu2021survey}. Factors such as varying illumination conditions and the poses of objects disproportionately impact the detection of small objects compared to larger ones. Moreover, an imbalance in class distribution can significantly hinder the performance of small-object detection models \cite{antony2024enhancing}.
With the evolution of deep learning methodologies, transfer learning and fine-tuning have become increasingly prevalent. Transfer learning involves updating only the final layers of a pre-trained model to adapt to new features, thereby preserving the generality of the learned features from the initial training while streamlining the adaptation process \cite{tan2018survey}. In contrast, fine-tuning adjusts all parameters of a pre-trained model, facilitating more extensive optimization for a new dataset by leveraging prior learned knowledge \cite{Yin.2017}. Both strategies offer distinct advantages in various deep learning scenarios.

Federated learning, a paradigm designed to enhance data privacy by sharing only model weights instead of raw data among clients, is emerging in the context of object detection \cite{fedavg}. Although federated learning has been explored in general object detection scenarios \cite{fedod_optimize}, \cite{Hegiste.}, \cite{fedvision}, \cite{fedod_visual}, its application to small object detection remains under-investigated. This gap in research is what this study seeks to address, inspired by findings from \cite{Hegiste.b} that demonstrate federated learning's potential to function as an ensemble technique, thereby improving model performance on distributed datasets.
In machine learning, regularization techniques are crucial for preventing overfitting, a common issue where a model learns the training data too well and fails to generalize to new data. Federated learning naturally incorporates regularization through the inherent heterogeneity and distribution of the training data across multiple devices, introducing variability and noise that can help regularize the model \cite{Aggarwal2015-DataMiningPatternsTechniques}, \cite{Hastie2009-ElementsOfStatisticalLearning}, \cite{Li2020-FedProx}, \cite{Acar2021-feddyn}. Additionally, the communication constraints inherent in federated learning can further regularize the model by limiting the amount of data shared between devices \cite{Wang2022-FederatedLearningIntermediateRepresentationRegularization}. This study aims to evaluate the performance of object detection models trained using various deep learning and federated learning techniques against an unseen test dataset, leveraging the regularization benefits of federated learning.

\FloatBarrier
\section{Methodology}
\subsection{Dataset} \label{dataset}

\begin{figure}
\centering
    \includegraphics[width= 0.38\textwidth]{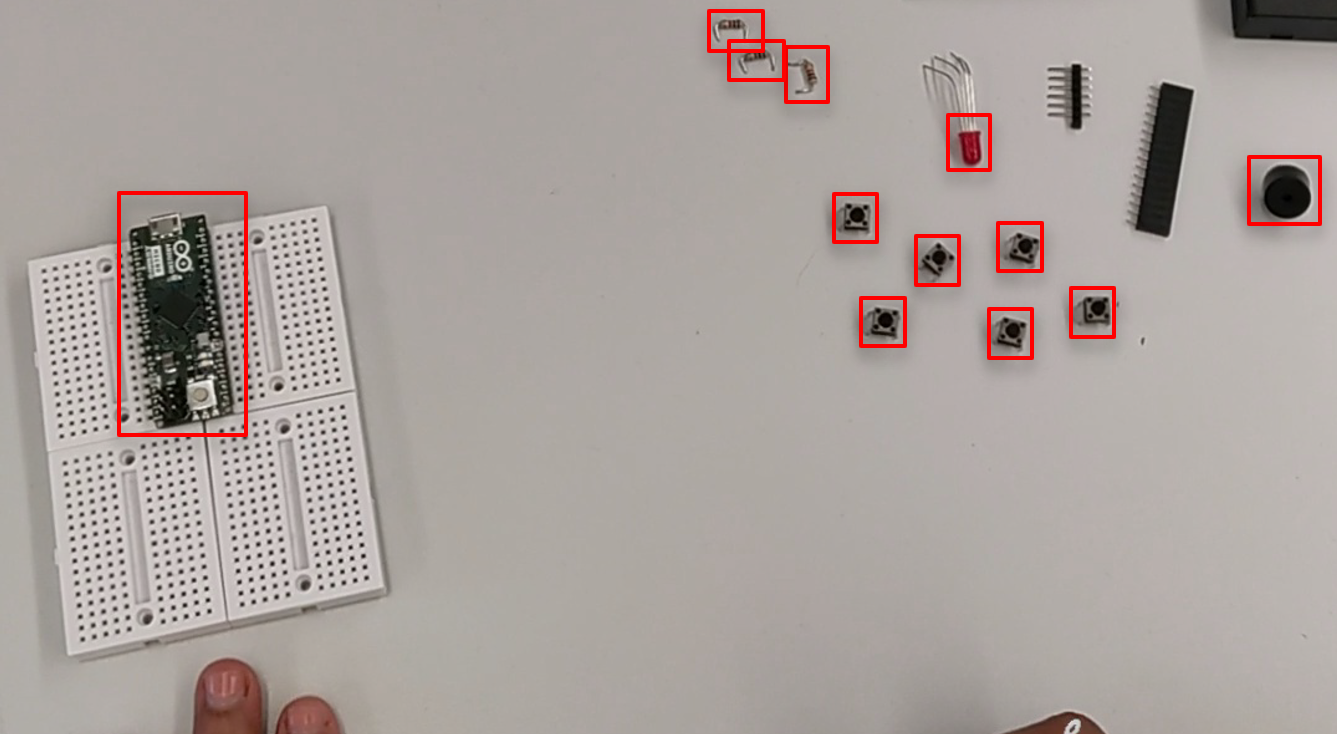}
    \caption{Image sample from the Real images dataset}
    \label{fig:real}
    
    \vspace{0.5cm}
    
    \includegraphics[width= 0.38\textwidth]{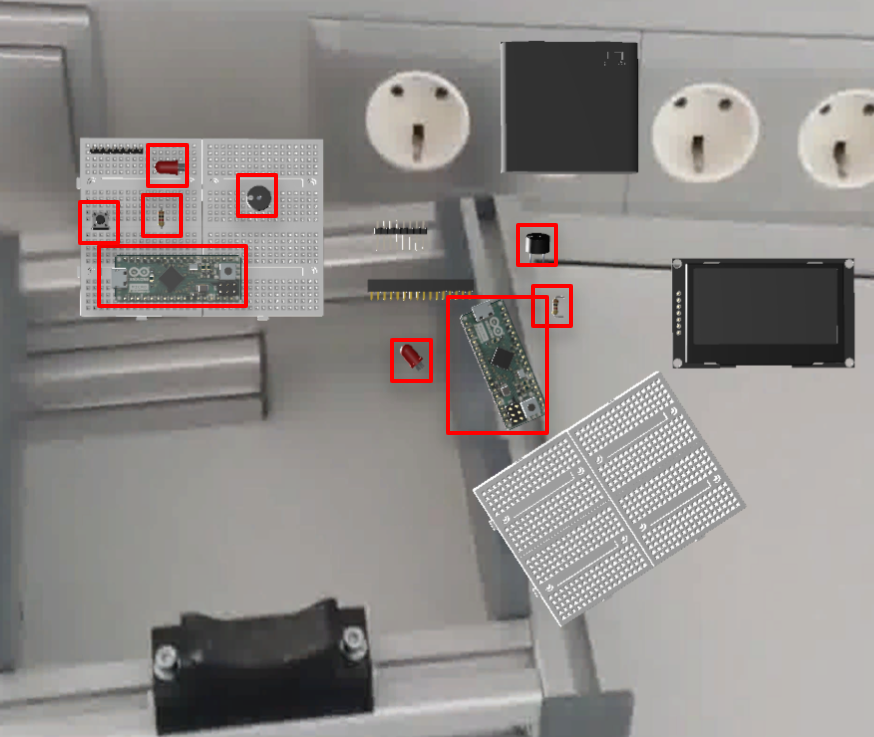}
    \caption{Image sample from the Synthetic images dataset}
    \label{fig:synthetic}
\end{figure}

The hybrid dataset includes real and synthetic images captured for both small and normal-sized objects from an assembly product, featuring: micro-Arduino, tactile buttons, resistors, LED lights, and buzzers mounted on a breadboard. The synthetic dataset is generated using CAD models in Unity3D scene, incorporating variable viewpoints, backgrounds, illuminations, camera to object distances, and object states (assembled and disassembled).
Fig. \ref{fig:real} and Fig. \ref{fig:synthetic} \footnote{Figures are cropped and enlarged for visual purpose} show samples from the real and synthetic datasets, respectively. The CAD models used are open source and simple, detailed in \cite{antony2024enhancing}.
The real dataset had a total of 3001 bounding boxes and synthetic with 2700 bounding boxes.

\begin{figure}
\centering
    \includegraphics[width= 0.38\textwidth]{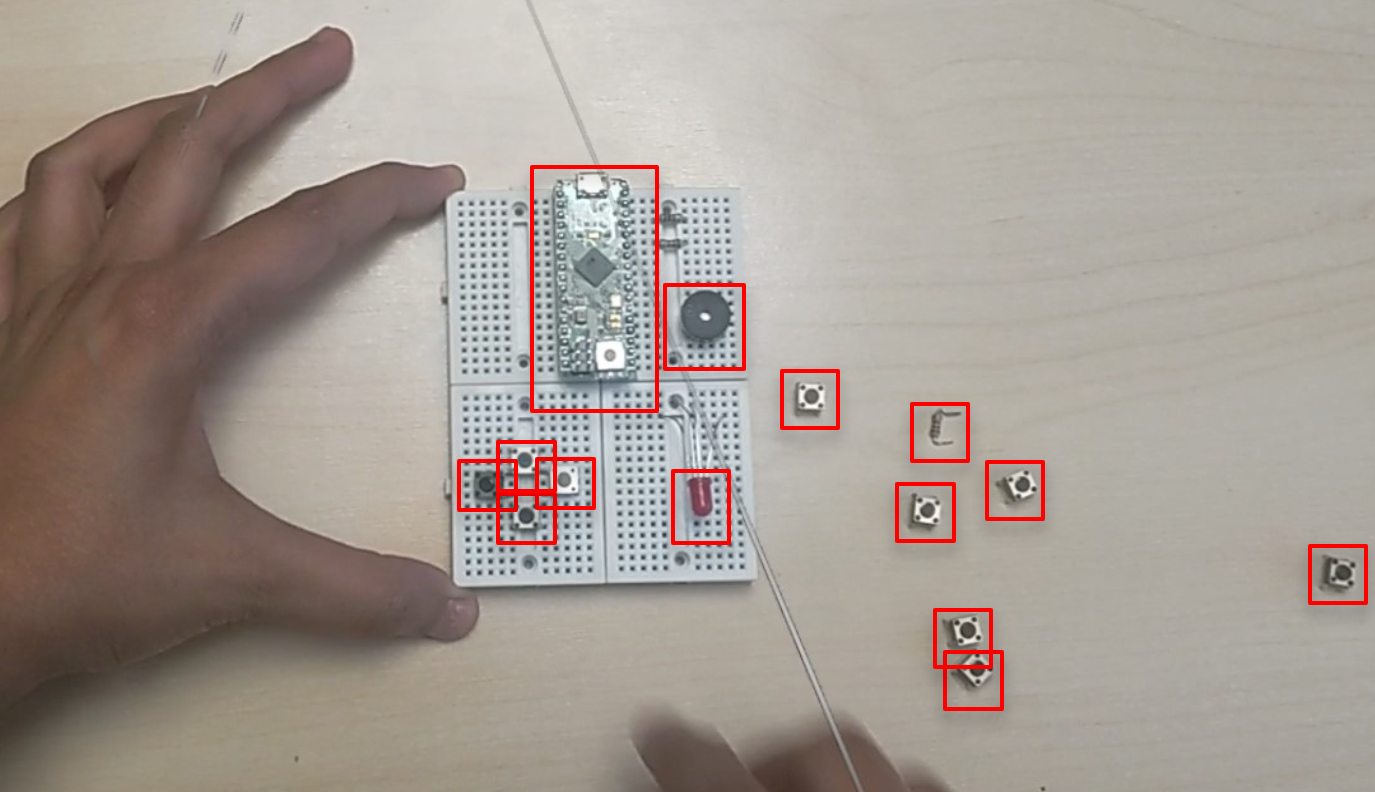}
    \caption{Test image having a different background, lighting and blur}
\label{fig:testdata}
\end{figure}

For testing the robustness of the model and facilitate better comparison, a new test dataset was created. This dataset includes 116 images with varied and cluttered backgrounds, lighting conditions, and motion blur. Sixteen images contain only backgrounds to assess false positives. The sample image is shown in Figure \ref{fig:testdata}. The remaining 100 images consist of a total of 1101 bounding boxes distributed across 5 classes.

\subsection{Algorithms}
This paper aims to compare the performance on a hybrid dataset using various algorithmic techniques, including:

\begin{enumerate}
    \item \textbf{Transfer Learning:} The model is initially trained with synthetic data and then applied to a real dataset.
    
    \item \textbf{Fine-tuning:} The model, initially trained with synthetic data, is further fine-tuned on a real dataset.
    
    \item \textbf{YOLOv5 Ensemble Technique:} An ensemble model is created by combining the output of two YOLOv5 models trained with a real and synthetic dataset.
    
    \item \textbf{Federated Learning:} The dataset is divided, with one client having a synthetic dataset and the other having a real dataset, employing federated learning to achieve a global model.
    
    \item \textbf{FedEnsemble Technique:} The centralized dataset is divided into three clients, and a global federated model is obtained through federated learning.
\end{enumerate}

These techniques are compared with centralized learning YOLOv5l model using a hybrid dataset, where an equivalent number of real and synthetic images are utilized, as detailed in \cite{antony2024enhancing}.

\section{Implementation} \label{implementation}

In line with the findings by \cite{antony2024enhancing}, the introduction of a synthetic dataset, even a simple one, during training significantly improves the mean Average Precision (mAP) of the model. For a small dataset of 300 real images, the optimal approach involved supplementing it with an equivalent number of synthetic data, ensuring class balancing, particularly for LEDs and Arduinos, which were present once per image.
While maintaining this dataset for training, a new test dataset was created to assess model performance in a different environment. This test dataset featured variations in backgrounds, lighting conditions, blur, and distance from the HoloLens. Annotations were applied to this new dataset, and the results obtained from all algorithms are detailed in Section \ref{resultanddiss}. 
All algorithms utilized the YOLOv5l model architecture and pre-trained YOLOv5l model weights as the starting point. The centralized training approaches incorporated early stopping to determine the epochs for optimal model weights. Specifically, the hybrid dataset was trained on using the YOLOv5l architecture, and the best weights were achieved after approximately 200 epochs.
\begin{figure}
\centering
    \includegraphics[width= 0.4\textwidth]{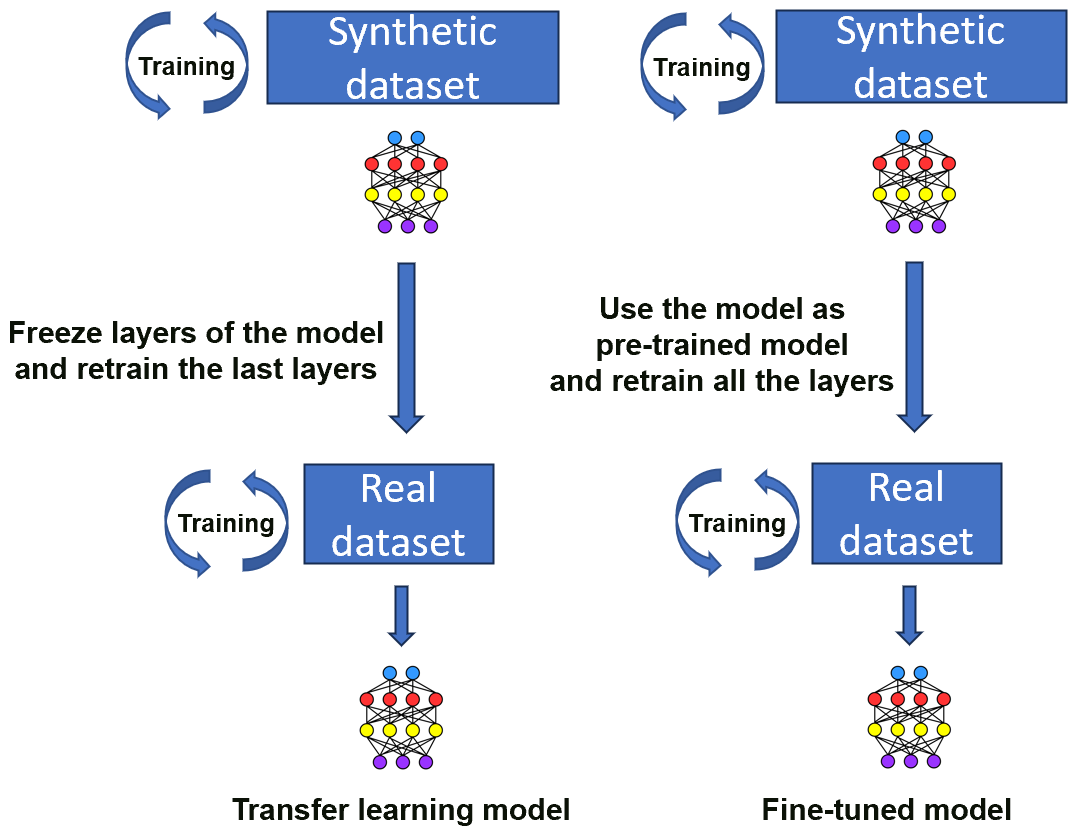}
    \caption{Transfer learning (left) and Fine-tune model (right) are first trained on synthetic and then the knowledge is transferred or fine-tuned to real image dataset}
    \label{fig:finetune}
    \vspace{0.3cm}
    \includegraphics[width= 0.4\textwidth]{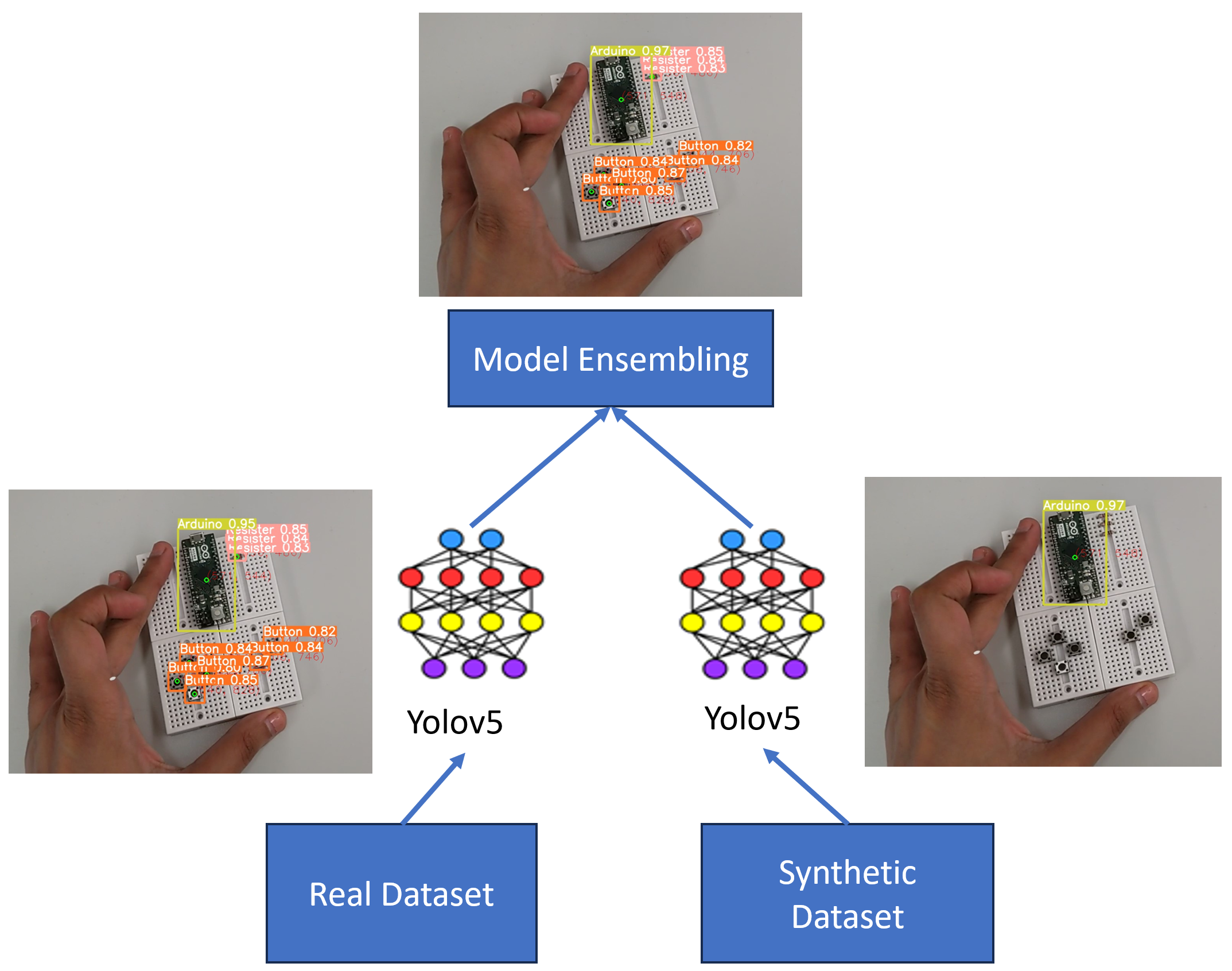}
    \caption{Model ensembling using individual Yolov5 as base models trained separately on synthetic and real data}
    \label{fig:ensemble}
\end{figure}

\subsection{Hyperparameters for Training Techniques}

\subsubsection{Fine-tuning}
The model was initially trained on a synthetic dataset of 300 images, achieving optimal performance on an NVIDIA A40 GPU cluster. In a second fine-tuning stage, the model was further trained on a real image dataset of 300 images. With early stopping, the optimal weights for the synthetic dataset were reached around 200 epochs. Subsequently, an additional 200 epochs were applied for fine-tuning these weights on the real dataset, as shown in Figure \ref{fig:finetune}.

\subsubsection{Transfer Learning}
The YOLOv5l model, fine-tuned on the synthetic dataset, underwent transfer learning. The core 10 layers of the trained model were frozen, and final layers were updated through training on the NVIDIA A40 GPU cluster using real-world scenario image data. With early stopping, the optimal weights for the synthetic dataset were reached around 200 epochs (as mentioned above). And the early stopping for transfer learning was reached around 150 epochs, as shown in Figure \ref{fig:finetune}.

\subsubsection{Model Ensembling}
The YOLOv5l model ensembling technique involved training two models separately on synthetic and real datasets. Inference on test images was performed by ensembling the results, showcasing improvements in ensemble inference, as shown in Figure \ref{fig:ensemble}.
These 2 models individually are also used for comparison, as shown in Table \ref{tab:old_test} and \ref{tab:new_test}.

\begin{figure} [h]
\centering
    \includegraphics[width= 0.38\textwidth]{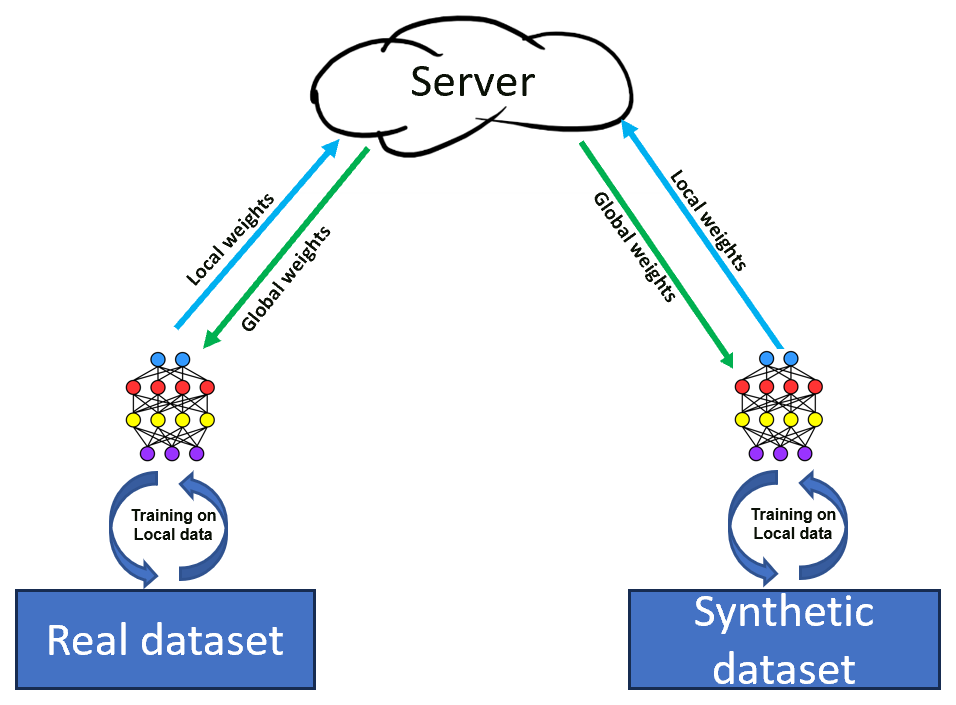}
    \caption{Federated learning with 2 clients with one having a real dataset and another synthetic dataset}
    \label{fig:fedl}
    \vspace{0.3cm}
    \includegraphics[width= 0.38\textwidth]{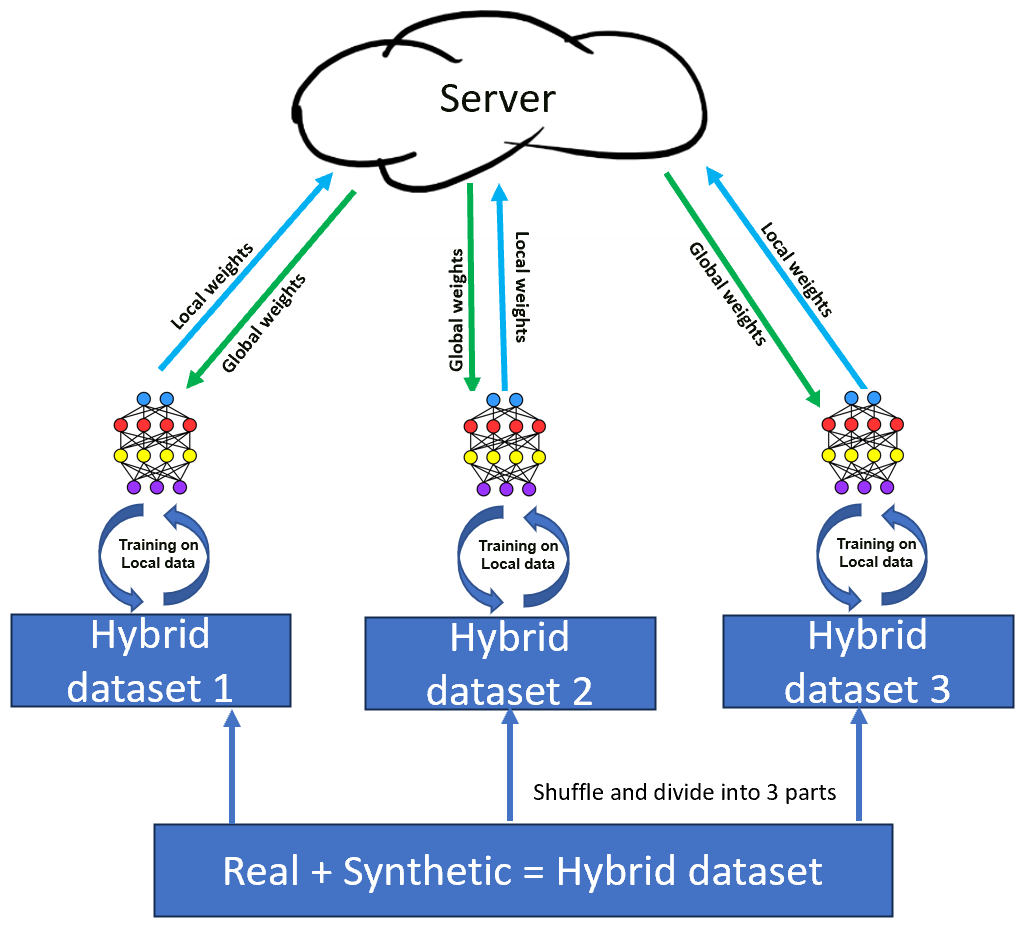}
    \caption{Federated ensemble (FedEnsemble) learning with 3 clients with each having a subset of the shuffled centralized hybrid dataset}
    \label{fig:fedensemble}
\end{figure}
\subsubsection{Federated Learning and FedEnsemble}
The hybrid learning dataset was partitioned into two clients for federated learning. Figure~\ref{fig:fedl} illustrates this process, where the first client received real images, and the second client received the synthetic dataset. The goal is to compare the global model obtained through federated learning with a model trained on a centralized hybrid dataset.
The FedAvg algorithm, ~\cite{fedavg}, was used for model aggregation. Various optimizers, including ADAM \cite{Kingma2014_adam}, ADAMW \cite{Loshchilov2017_adamw}, SGD \cite{Rumelhart1986_sgd}, and SGD with momentum \cite{sgdm}, were explored.
The object detection model used was YOLOv5l for consistency across experiments. Optimal global model weights were attained using SGD with momentum, 15 local epochs, and 10 communication rounds.

For the FedEnsemble method~\cite{Hegiste.b}, the centralized hybrid dataset was divided into three client datasets, as shown in Figure~\ref{fig:fedensemble}. Using YOLOv5l and a similar setup as federated learning with the FedAvg algorithm, the global model was achieved with 15 local epochs and 10 communication rounds. FedEnsemble demonstrated increased robustness compared to centrally trained models on objects in an unseen environment~\cite{Hegiste.b}. 

The best models were selected after extensive epochs and communication rounds combinations, and were tested on 100 images from the same dataset distribution as the training dataset and an additional 116 test images, as mentioned in Subsection~\ref{dataset}. Detailed results are presented in Section~\ref{resultanddiss}. All architectures had the same optimizer, SGD with momentum=0.937, batch size=8, IoU training threshold=0.20 image size=1080, weight decay=0.0005, lr=0.01 etc. as mentioned Yolov5 default training setting.

\FloatBarrier
\section{Results and Discussion} \label{resultanddiss}

\begin{table*}[h]
\centering
\begin{tabular}{|l|l|l|l|l|l|l|l|}
\hline
Algorithm & AP & AP50 & AP75 & APsmall & APmedium & APlarge & mAP \\ \hline\hline
FedEnsemble model & 0.6530 & 0.9718 & 0.7643 & 0.6212 & 0.6349 & 0.4101 & 0.9749 \\ \hline
Federated learning & 0.5723 & 0.9586 & 0.5825 & 0.5011 & 0.5702 & 0.3945 & 0.9660 \\ \hline
Transferlearning & 0.6635 & 0.9725 & 0.7763 & 0.6092 & 0.6453 & 0.4142 & 0.9755 \\ \hline
Fine-tune model & \textbf{0.6680} & \textbf{0.9743} & 0.7124 & \textbf{0.6671} & \textbf{0.6561} & \textbf{0.4247} & 0.9774 \\ \hline
Hybrid Centralized learning & 0.6643 & 0.9702 & 0.7455 & 0.6318 & 0.6399 & 0.4240 & \textbf{0.9783} \\ \hline
Real Centralized learning & 0.6676 & 0.9663 & \textbf{0.7820} & 0.6598 & 0.6472 & 0.4107 & 0.9743 \\ \hline
Synthetic Centralized learning & 0.1509 & 0.3595 & 0.1025 & 0.0902 & 0.1890 & 0.0946 & 0.3663 \\ \hline
YOLO Ensemble & 0.5314 & 0.9171 & 0.5470 & 0.5731 & 0.5232 & 0.3036 & 0.9188 \\ \hline
\end{tabular}
\vspace{0.1cm}
\caption{Comparison of all models based on COCO metrics and mAP on test dataset which was captured in the same environment as the training dataset}
\label{tab:old_test}

\begin{tabular}{|l|l|l|l|l|l|l|l|}
\hline
Algorithm & AP & AP50 & AP75 & APsmall & APmedium & APlarge & mAP \\ \hline\hline
FedEnsemble model & 0.4375 & 0.8193 & 0.4108 & 0.2882 & 0.4188 & 0.7480 & 0.8212 \\ \hline
Federated learning & \textbf{0.4807} & \textbf{0.8600} & \textbf{0.4729} & \textbf{0.2969} & \textbf{0.4555} & 0.7580 & \textbf{0.8638} \\ \hline
Transferlearning & 0.3812 & 0.7544 & 0.3417 & 0.2373 & 0.3272 & 0.7245 & 0.7681 \\ \hline
Fine-tune model & 0.4239 & 0.7726 & 0.4142 & 0.2912 & 0.3606 & \textbf{0.7837} & 0.7856 \\ \hline
Hybrid Centralized learning & 0.4432 & 0.7776 & 0.4726 & 0.2843 & 0.3801 & 0.7531 & 0.7865 \\ \hline
Real Centralized learning & 0.3955 & 0.7481 & 0.4041 & 0.2556 & 0.3559 & 0.7322 & 0.7638 \\ \hline
Synthetic Centralized learning & 0.2096 & 0.4080 & 0.1841 & 0.0875 & 0.2445 & 0.3828 & 0.4066 \\ \hline
YOLO Ensemble & 0.3693 & 0.7566 & 0.3230 & 0.2709 & 0.3451 & 0.6422 & 0.7664 \\ \hline
\end{tabular}
\vspace{0.1cm}
\caption{Comparison of all models based on COCO metrics and mAP on test dataset which was captured in an entirely different environment than the training dataset}
\label{tab:new_test}
\end{table*}

This section showcases a comparison of results from all the algorithms on two test datasets. The first dataset (referred to as Testset1) is similar to the background and setup shown in image \ref{fig:real}, and the second dataset (referred to as Testset2) features unseen environment parameters, as shown in figure \ref{fig:testdata}.
Table \ref{tab:old_test} presents results based on Testset1, while Table \ref{tab:new_test} presents results based on Testset2 in COCO metrics \cite{cocometrics}, which includes IoU-Aware (Intersection over Union) and object size-relevant metrics. The last column denotes mAP (mean Average Precision) at an IoU threshold of 0.5 in terms of PASCAL metrics, which is a single IoU threshold metric. Tables \ref{tab:old_test} and \ref{tab:new_test} include two additional tests beyond those mentioned in section \ref{implementation}.
The 'Real Centralized learning' test uses only the 300 real images dataset to train the YOLOv5 model for 200 epochs, and the resulting best model weights are used to test both Testset1 and Testset2. Similarly, the 'Synthetic Centralized learning' model is trained using only the 300 synthetic images from the hybrid dataset and is tested on both datasets.

Starting with Table \ref{tab:old_test}, we observe that most models achieve an mAP of more than 95\%, except for the model trained using only the synthetic dataset and the YOLOv5 ensemble model (which uses this synthetic dataset model as one of its two models). These two models exhibit a lower mAP, and the average precision for small objects is also lower. The model trained solely on the real dataset of 300 images performs well in all AP metrics.
Federated learning and FedEnsemble learning techniques perform on par with the centrally trained model but do not achieve better results. Transfer learning, fine-tune learning, and the model trained directly on the hybrid dataset also perform well on the test dataset, achieving APsmall greater than 60\% and mAP above 97\%. The fine-tune model stands out among all the algorithms, performing well in terms of all AP metrics. However, it's important to note that the test images are from the same sample set distribution as the training dataset (500 images were captured, of which 300 were used for training, 100 for validation, and 100 for testing, as shown in figure \ref{fig:real}). Hence, it is yet to be confirmed if the models really perform well or are overfitted to the training dataset samples.

To test the real accuracy of the models, all the models were tested on a newly created test dataset (as mentioned in section \ref{dataset}, figure \ref{fig:testdata}).
Table \ref{tab:new_test} reveals that the 'Real Centralized learning' model and the transfer learning model achieve an mAP of around 76\%, with APsmall at 25\% and 23\%, respectively. The fine-tuned model and hybrid dataset model only slightly outperform them, achieving around 75\% mAP and 28\% APsmall metrics. This strongly suggests that the centrally trained dataset models tend to overfit to the training dataset, especially the model trained only using 300 real images. The synthetic dataset model underperforms here, as does the YOLOv5 ensemble model.

The results for federated learning from Table \ref{tab:new_test} show that federated learning used to create a hybrid model by training two clients, each having only real and synthetic datasets respectively, produced the best results. The results are followed by the FedEnsemble learning, which also succeeded in producing a robust model that performs well on unseen environment datasets by using the same dataset used for centralized training but divided into different subsets as clients' datasets.
The results of the global federated model are 8\% better in terms of PASCAL metrics mAP (mean Average Precision) when compared to the baseline centrally trained model. In terms of COCO metrics \cite{cocometrics}, federated learning outperforms the baseline model in all AP (Average Precision) types, as seen in the Table. Federated learning is seen to be performing better than all the other algorithms, followed by the FedEnsemble technique, which also performs better on unseen test data within a different environment (As shown in Figure \ref{fig:hybrid_output} and \ref{fig:fedens_output} for centralized and FedEnsemble model respectively).

\begin{figure} [h]
\centering
    \includegraphics[width= 0.3\textwidth]{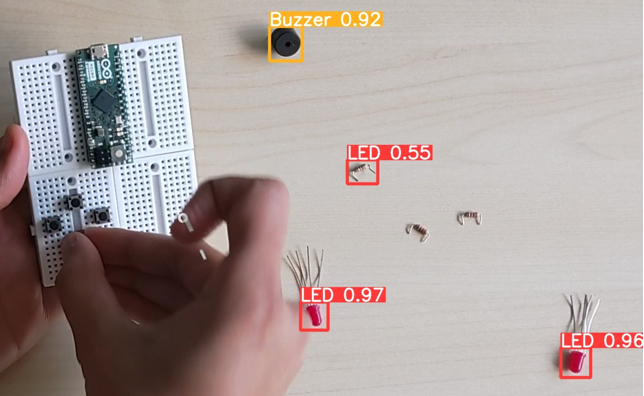}
    \caption{Output of \textbf{YOLOv5l model trained using a Hybrid centralized dataset} on test dataset with unseen environment}
    \label{fig:hybrid_output}
    \vspace{0.3cm}
    \includegraphics[width= 0.3\textwidth]{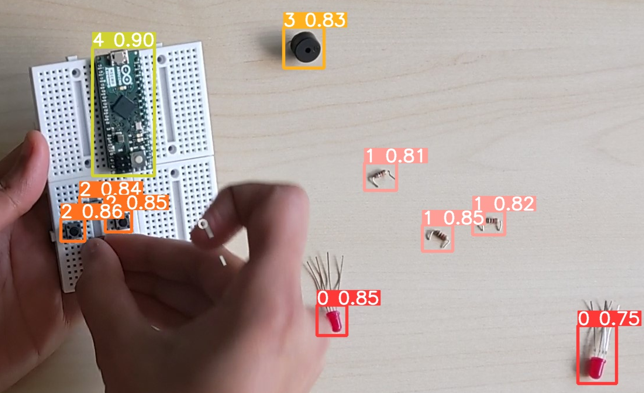}
    \caption{Output of \textbf{Global FedEnsemble model} on test dataset with unseen environment}
    \label{fig:fedens_output}
\end{figure}
The idea of testing models trained with only synthetic or real datasets is to showcase that clients' models trained with local datasets performed subpar when similar objects were placed in a new unseen environment condition. However, using similar datasets with the help of federated learning, clients can achieve a robust global federated model without sharing any raw private data (Refer Appendix for further details).

\section{Conclusion}
In conclusion, this paper conducted a comparative analysis between federated learning and centralized models trained on a hybrid dataset. The centralized models, trained with a limited dataset of 300 images and 300 synthetic images generated with class balancing, exhibited signs of overfitting. While they demonstrated strong performance on images from a similar distribution as the training dataset, their efficacy declined significantly when confronted with a test dataset featuring novel environmental parameters, such as distinct backgrounds, lighting conditions, blur, and camera angles.
Contrastingly, the federated learning global models for object detection showcased robust performance on this diverse test dataset. The results underscore the resilience of the federated global model and the FedEnsemble model, both of which demonstrated superior performance and did not exhibit the false positives observed in centralized models when tested on the challenging dataset. This suggests the potential of federated learning for enhancing model generalization across varying environmental conditions.

The main outcome of these results also portray that Federated learning and FedEnsemble act as regularization due to communication constraints and data heterogeneity, producing a robust model that does not overfit to a single dataset, unlike the models trained with a centralized dataset approach in the case of small datasets.

\section{Future Work}
Use of federated learning should be promoted for a robust model creation for machine learning use cases.
Another perspective is that FedEnsemble can be used as an ensemble technique even if a centralized dataset is available. 
A proper synthetic dataset can further benefit your machine learning model in manufacturing scenarios where the dataset samples are limited.
This work can be further extended by experimenting with different manufacturing use cases and generic datasets where datasets are small with repetitive samples and comparing the federated learning models with centralized trained models.

\bibliographystyle{IEEEtran.bst}
\bibliography{lit}

\appendices 

\section{Output of Models on Test Dataset} \label{FirstAppendix}

\begin{figure*} [h]
\centering
    \includegraphics[width= 0.95\textwidth]{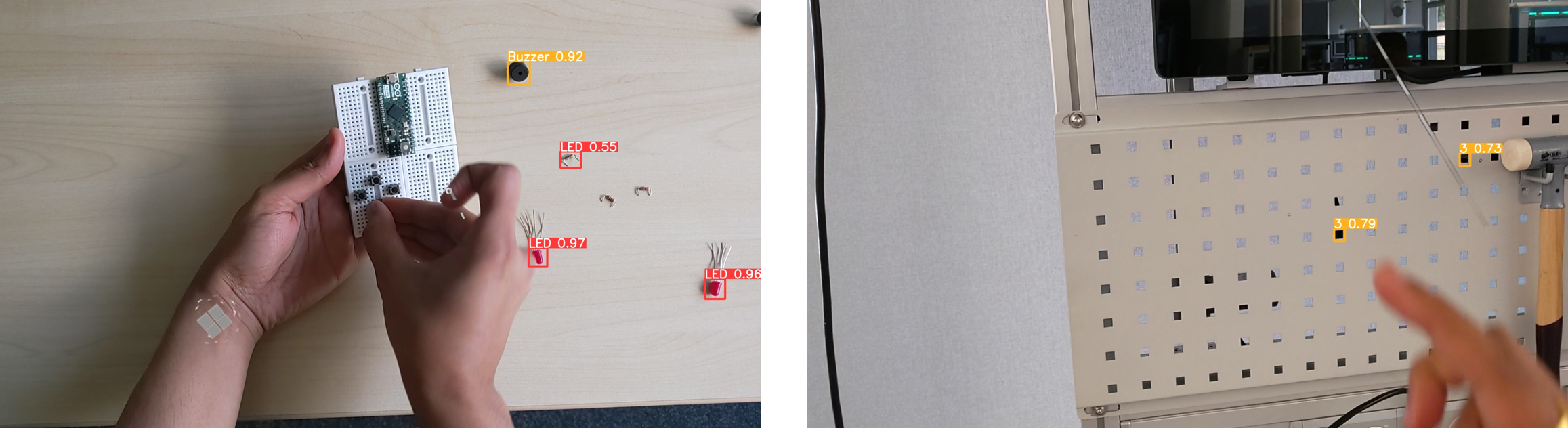}
    \caption{Output of \textbf{YOLOv5l model trained using a Hybrid centralized dataset} on test dataset with unseen environment (left) and on image with just background (right)}
    \label{fig:hybrid_output}
    \vspace{0.3cm}
    \includegraphics[width= 0.95\textwidth]{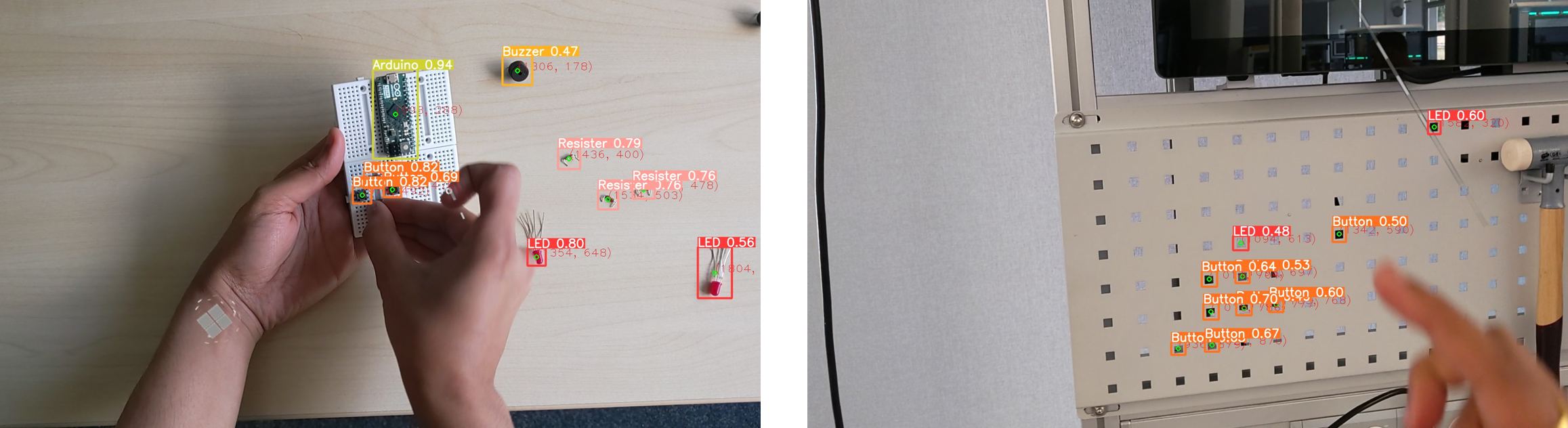}
    \caption{Output of\textbf{ Transfer learning model} on test dataset with unseen environment (left) and on image with just background (right)}
    \label{fig:transfer_output}
    \vspace{0.3cm}
    \includegraphics[width= 0.95\textwidth]{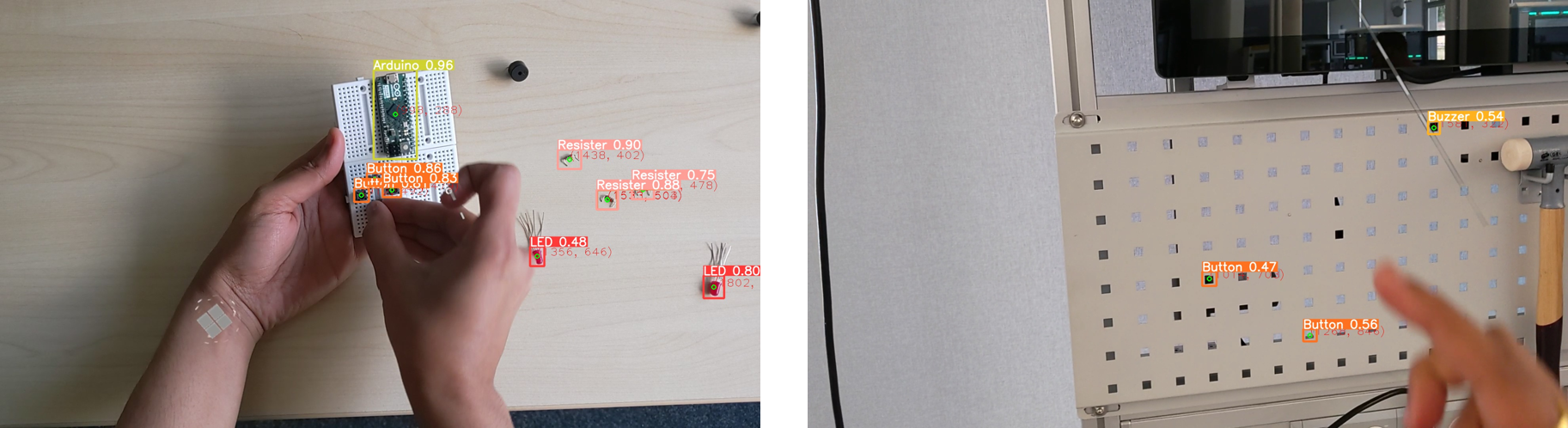}
    \caption{Output of \textbf{Fine-tuned model} on test dataset with unseen environment (left) and on image with just background (right)}
    \label{fig:finetune_output}
    \vspace{0.3cm}
    \includegraphics[width= 0.95\textwidth]{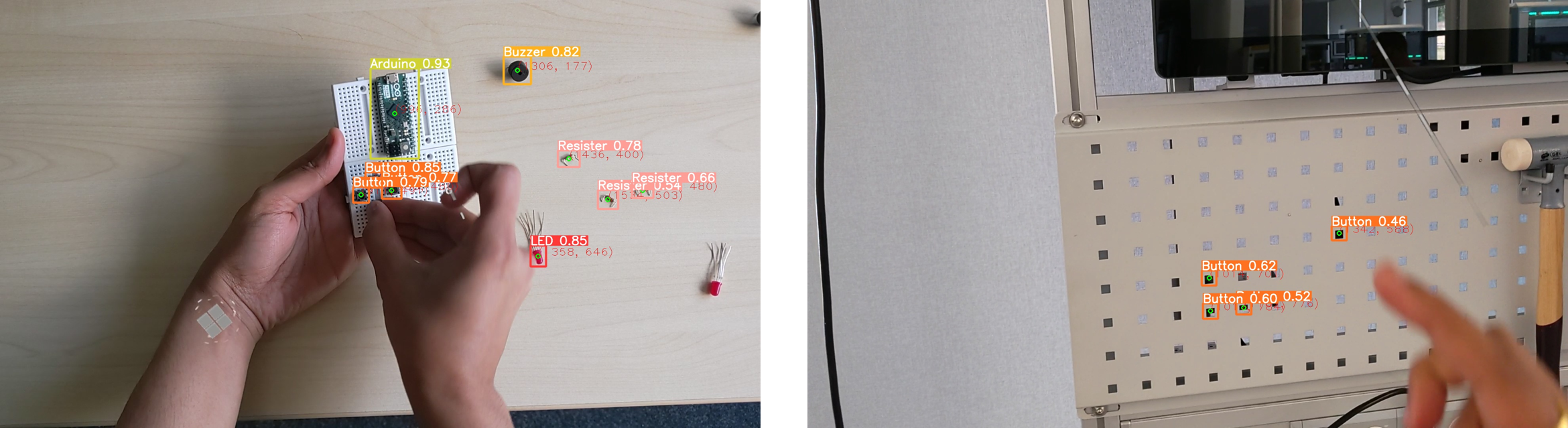}
    \caption{Output of \textbf{YOLOv5l model trained using only the real images} on test dataset with unseen environment (left) and on image with just background (right)}
    \label{fig:real_output}
\end{figure*}

\begin{figure*} [h]
\centering

    \includegraphics[width= 0.95\textwidth]{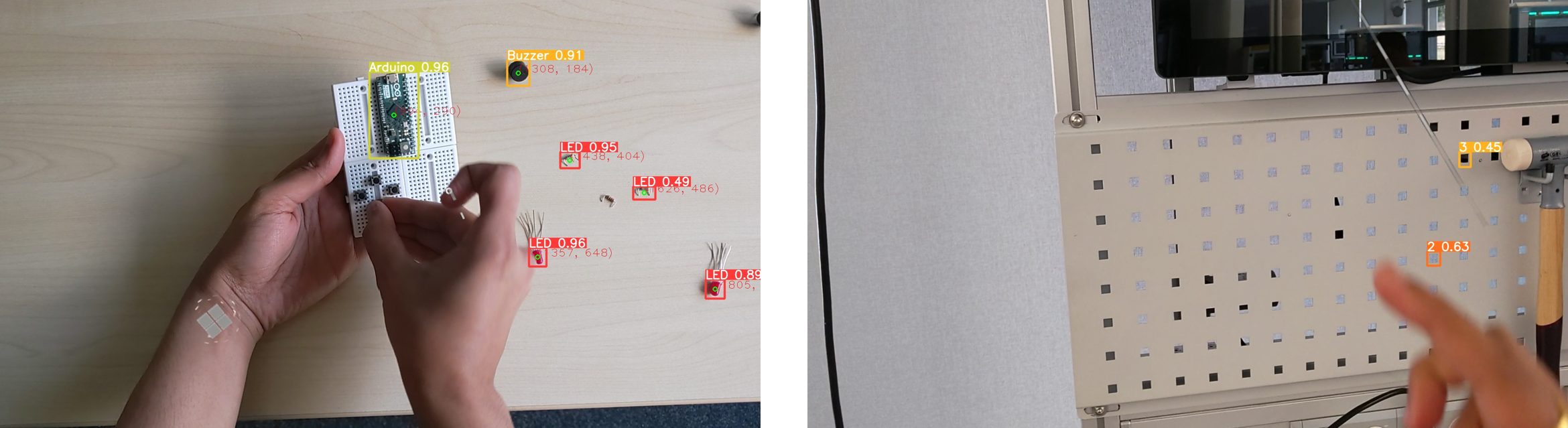}
        \caption{Output of \textbf{YOLOv5l model trained using only the synthetic dataset} on test dataset with unseen environment (left) and on image with just background (right)}
        \label{fig:syn_output}
        \vspace{0.3cm}
        \includegraphics[width= 0.95\textwidth]{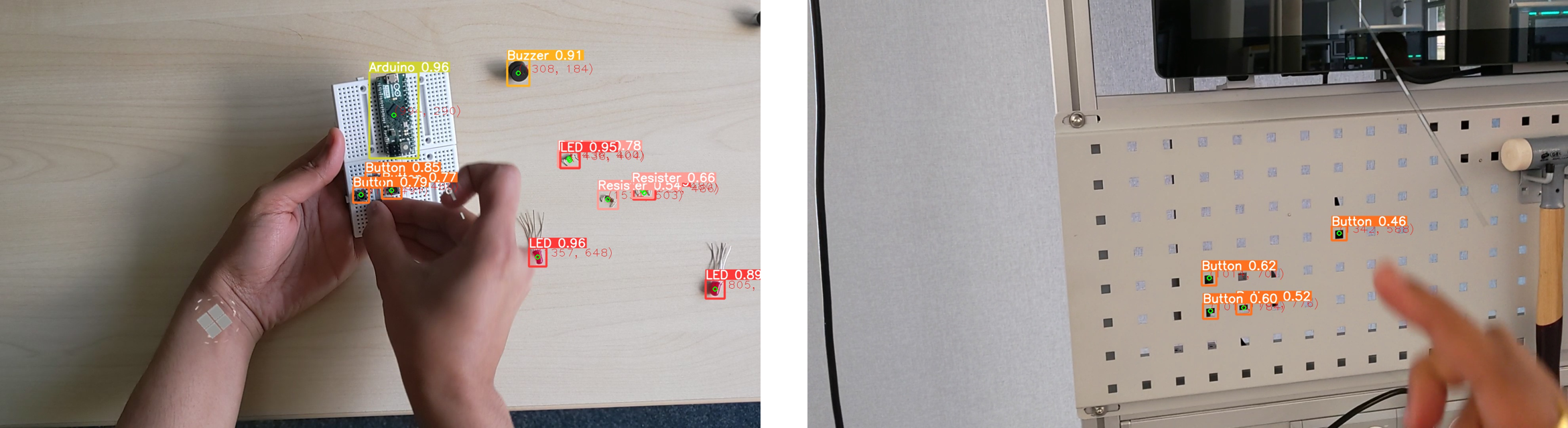}
        \caption{Output of \textbf{YOLOv5l real and synthetic ensemble model} on test dataset with unseen environment (left) and on image with just background (right)}
        \label{fig:ensemble_output}
        \vspace{0.3cm}
        \includegraphics[width= 0.95\textwidth]{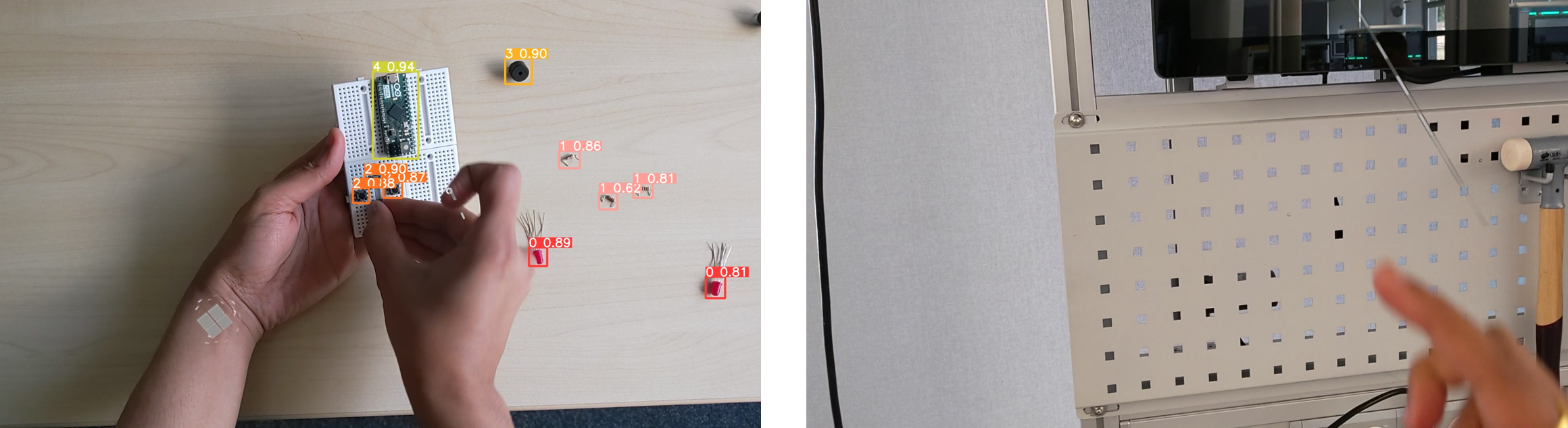}
        \caption{Output of \textbf{Global Federated model} on test dataset with unseen environment (left) and on image with just background (right)}
        \label{fig:fedl_output}
        \vspace{0.3cm}
        \includegraphics[width= 0.95\textwidth]{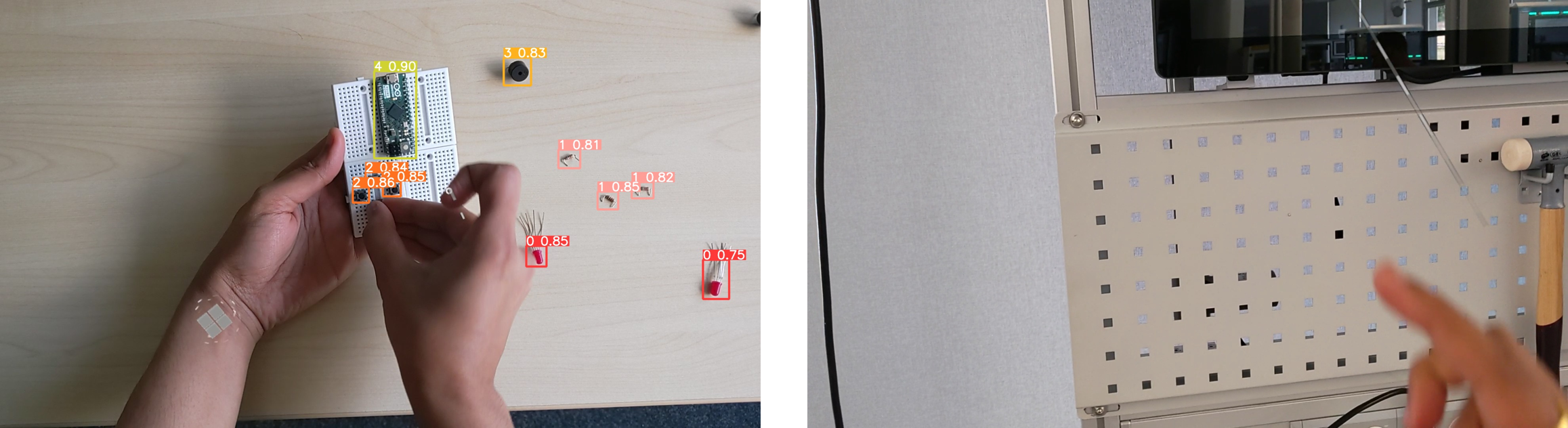}
        \caption{Output of \textbf{Global FedEnsemble model} on test dataset with unseen environment (left) and on image with just background (right)}
        \label{fig:fedens_output}

\end{figure*}

The output results from all the models on unseen test images can be referred to below. In Figures \ref{fig:hybrid_output}, \ref{fig:transfer_output}, \ref{fig:finetune_output}, \ref{fig:real_output}, \ref{fig:syn_output}, \ref{fig:ensemble_output}, \ref{fig:fedl_output}, and \ref{fig:fedens_output}, the left image is a test image consisting of all the objects in a different environment and lighting condition. We can see that the best-performing model outputs on these images are both the federated learning models.
Moving to the image on the right of all these figures, it is just a background image with no target objects in it, to test the false positives produced by the models. However, this background consists of a holed grill, which also looks very similar to a button or buzzer, and hence there are false positives in the output of all the models except for federated learning models (Figure \ref{fig:fedl_output} and \ref{fig:fedens_output}).

This proves that federated learning models are robust in variable testing environments, specially performs better than traditional techniques when the testing environments such as lighting conditions, blur, background, camera angle and distance from the target objects, etc.are not similar to the training image environments.

\end{document}